\documentclass[a4paper]{article}

\usepackage{amsmath}
\usepackage{color, multirow}
\usepackage{microtype}
\usepackage{graphicx}
\usepackage{booktabs} 
\usepackage{caption} 
\usepackage{makecell}
\usepackage{enumitem}
\usepackage{subfiles}
\usepackage{amsmath}
\usepackage{nicefrac}   
\usepackage{array}
\usepackage{bbm}
\usepackage{tikz}
\usepackage{adjustbox}

\newcommand*\question[1]{\textbf{Q{#1}}}
\usepackage{INTERSPEECH2020}

\title{Semantic Complexity in End-to-End Spoken Language Understanding}
\name{Joseph P. McKenna*\thanks{* These authors contributed equally.}, Samridhi Choudhary*, Michael Saxon*, Grant P. Strimel,\\ Athanasios Mouchtaris}
\address{
	Alexa Machine Learning, Amazon, USA
	\email{\{joemcke, samridhc, michsaxo, gsstrime, mouchta\}@amazon.com}}

\begin{document}

\maketitle
\begin{abstract}
 End-to-end spoken language understanding (SLU) models are a class of model
 architectures that predict semantics directly from speech. Because of their
 input and output types, we refer to them as speech-to-interpretation (STI)
 models. Previous works have successfully applied STI models to targeted use cases,
 such as recognizing home automation commands, however no study has yet addressed
 how these models generalize to broader use cases. In this work, we analyze the
 relationship between the performance of STI models and the difficulty of the use
 case to which they are applied. We introduce empirical measures of dataset {\it semantic
 	complexity} to quantify the difficulty of the SLU tasks. We show that
 near-perfect performance metrics for STI models reported in the literature were obtained
 with datasets that have low semantic complexity values. We perform experiments
 where we vary the semantic complexity of a large, proprietary dataset and show
 that STI model performance correlates with our semantic complexity measures,
 such that performance increases as complexity values decrease. Our results show that it is important to contextualize an STI model's performance with the complexity values of its training dataset to reveal the scope of its applicability.
\end{abstract}
\noindent\textbf{Index Terms}: spoken language understanding, semantic complexity, speech-to-interpretation

\section{Introduction} \label{sec:intro} Spoken language understanding (SLU) is
the task of inferring the semantics of user-spoken utterances. Typically, SLU is
performed by populating an utterance \textit{interpretation} with the results of
three sub-tasks: \textit{domain classification (DC)}, \textit{intent
	classification (IC)}, and \textit{named entity recognition
	(NER)}~\cite{tur2011spoken}. The conventional approach to SLU uses two
distinct components to sequentially process a spoken utterance: an automatic
speech recognition (ASR) model that transcribes the speech to a text transcript,
followed by a natural language understanding (NLU) model that predicts the
domain, intent and entities given the transcript~\cite{lee2010recent}. Recent
applications of deep learning approaches to both ASR \cite{hinton2012deep,
	graves2013speech, bahdanau2016end} and NLU \cite{xu2014contextual,
	ravuri2015recurrent, sarikaya2014application} have improved the accuracy and
efficiency of SLU systems and driven the commercial success of voice
assistants such as Amazon Alexa and Google Assistant. However, the modular
design of traditional SLU systems admits a major drawback. The two component
models (ASR and NLU) are trained independently with separate objectives. Errors
encountered in the training of either model do not inform the other. As a
result, ASR errors produced during inference might lead to incorrect NLU
predictions since the ASR errors are absent from the clean transcripts on which
NLU models are trained.

An increasingly popular approach to address this problem is to employ models
that predict SLU output directly from a speech signal
input~\cite{lugosch2019speech,haghani2018audio,chen2018spoken,serdyuk2018towards}.
We refer to this class of SLU models as speech-to-interpretation (STI). STI
models can be trained with a single optimization objective to recognize the
semantics of a spoken utterance; some studies report better performance
by pretraining initial layers of the model on transcribed speech followed by
fine-tuning with semantic
targets~\cite{lugosch2019speech,chen2018spoken}. The semantic target types,
predicted by previously published models, vary from just the
intent~\cite{lugosch2019speech, chen2018spoken} to the full SLU
interpretation~\cite{haghani2018audio}. A majority of these works report
near-perfect performance metrics for DC or IC on independently collected
datasets. For example, IC accuracy values of $98.8\%$~\cite{lugosch2019speech}
and $98.1\%$~\cite{chen2018spoken}, DC accuracy of
$97.2\%$~\cite{serdyuk2018towards}, and F1-scores of $96.7\%$ and $95.8\%$ for
DC and IC respectively~\cite{haghani2018audio} have been reported.

While these studies demonstrate successful applications of STI architectures,
their results are often benchmarked on datasets with limited complexity. Most
publicly available SLU datasets used to train STI models contain only a moderate
number of unique utterances and intent
classes~\cite{lugosch2019speech,saade2018spoken,picovoice}. For example, the
Fluent Speech dataset ~\cite{lugosch2019speech} only contains 248 unique
utterance phrasings from one domain. Similarly, transcriptions within some
intent classes in the dataset used in \cite{chen2018spoken} are keywords such as
`true' and `false'.

As a result, the question of how well STI models generalize to broader use cases
remains unexplored.

In this work, we study the relationship between the performance of STI models
and the inherent difficulty of their use cases. We pose
the following research questions:
\begin{enumerate}[label=\protect\question{\arabic*}]\addtolength\itemsep{-1.3mm}
	\item Can the semantic complexity of a dataset be quantified to understand the
	difficulty of an STI task?
	\item How does varying the semantic complexity measure of a dataset affect the
	STI model performance?
\end{enumerate}
We propose several empirical measures of the {\it semantic complexity} of an SLU
dataset (\question{1}). These measures help quantify the difficulty of an SLU
task underlying a particular use case. We ground the near-perfect performance
metrics reported in literature with complexity values for their training
datasets and show that they have low semantic complexity. We also perform
experiments to study the relationship between our proposed complexity measures
and the performance of STI models (\question{2}). We advocate for reporting the
complexity values of SLU datasets along with the performance of STI models to
help uncover the applicability of the proposed architectures. In the following sections, we
describe our semantic complexity measures, report experimental results, and
conclude with a discussion and suggestions for future work in this area.

\section{Semantic Complexity} \label{sec:complex} Our first research question
(\question{1}) poses if quantifying the semantic complexity of an SLU dataset
can indicate the difficulty of the underlying SLU task. We propose several
data-driven measures of semantic complexity that are computed using the SLU
dataset transcriptions. These measures can be grouped into two broad categories:
{\it lexical} measures and {\it geometric} measures. The major difference
between these groups is the representation used for the transcriptions. The
lexical measures use common lexical features, like n-grams, whereas the
geometric measures use vectors encodings to represent the transcription text. In
this section we define each measure and in Section \ref{sec:exp} we show that
STI model performance is correlated with semantic complexity computed using
these measures.

\subsection{Lexical Measures}
The lexical measures that we consider are {\it vocabulary size}, {\it number of
	unique transcripts}, and {\it n-gram entropy}. The vocabulary size measures
the number of unique words in the dataset transcripts.\\

\noindent \textbf{n-Gram Entropy}: The n-gram entropy measures the randomness of
the dataset transcripts over its constituent n-grams, $\mathcal{N}$. It is
computed by the equation:
\begin{equation*}
H=-\sum_{x\in \mathcal{N}^*} p(x)\log_2  p(x)
\end{equation*}
where $\mathcal{N}^*$ is the set of unique n-grams in the dataset and $p(x)$ is
the probability of n-gram $x$ occurring in $\mathcal{N}$. A dataset that
contains only a single unique n-gram has an entropy of 0, whereas a dataset with
no repeated n-grams has an entropy of $\log_2 |\mathcal{N}|$ (uniform
distribution over $\mathcal{N}$). Larger values of n-gram entropy represent
higher randomness and variety in the utterance patterns, indicating larger
semantic complexity.

\subsection{Geometric Measures}
For this class of measures, the dataset transcripts are first encoded in a
high-dimensional Euclidean space, then subsequent analyses are performed. We
propose two measures under this category: {\it minimum spanning tree (MST)
	complexity} and {\it adjusted Rand index (ARI) complexity}. First, we explain
how to compute these two measures given any transcript encoding. Second, we
describe the encoding methods that we used. These measures are computed with the
assumption that each example in the dataset has a
semantic label, such as an intent label.\\

\begin{figure}[!t]
	\centering \includegraphics[width=\linewidth]{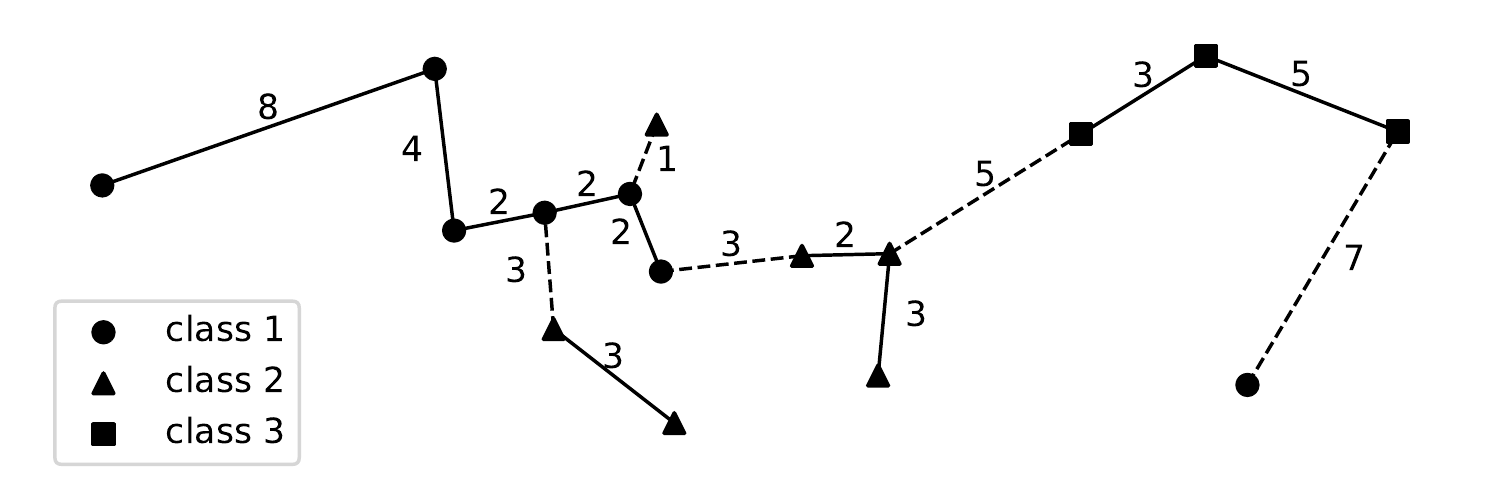}
	\caption{Depiction of minimum spanning tree complexity for a toy dataset. The
		$5$ edges in the MST that are inter-class (dashed lines) have weight $19$
		out of the total weight $W=53$, so the complexity is $19/53\approx0.36$.}
	\label{fig:mst}
\end{figure}

\noindent \textbf{Minimum Spanning Tree Complexity (MST)}: We hypothesize that
examples with similar transcript encodings yet different semantic labels will
confuse an STI model. To capture this notion, we compute a minimum spanning tree
(MST) of the transcript encodings. The MST is computed over a complete graph
where the vertices correspond to the transcripts and the edges are weighted by
the distance between their encodings. The edges of this MST connect pairs of
examples with similar transcript encodings, though some pairs will have
different semantic labels. We define the MST complexity measure as the
cumulative weight of MST edges that connect examples with different labels using
the expression:

\begin{equation*}\label{eq:mst}
{1 \over W} \sum_{(u,v)\in \text{MST}} w_{uv}~\mathbbm{1}{\left\{l(u) \neq l(v)\right\}}
\end{equation*}
where $l(u)$ and $l(v)$ are the labels of examples with transcript encodings $u$
and $v$, $w_{uv}$ is the weight of the edge between $u$ and $v$, and
$\textstyle{W=\sum_{(u,v)\in\text{MST}}w_{uv}}$ is the total edge weight of the
MST (see Figure~\ref{fig:mst}). The MST complexity ranges from $0$ to $1$, with
$1$ indicating the highest semantic
complexity.\\

\noindent \textbf{Adjusted Rand Index Complexity (ARI)}: As a complementary view
to the MST complexity, we hypothesize that semantic classes whose transcripts do
not cluster together in the encoding space could be confusing to an STI model.
To capture this notion, we first perform an agglomerative clustering\footnote{We
	used the complete-linkage criteria with the cosine distance.} on the
transcript encodings. After performing the clustering, each transcript has a
semantic class label assigned by the data annotator and a cluster label assigned
by the clustering algorithm. The adjusted Rand index
(ARI)~\cite{hubert1985comparing} quantifies the agreement between the two
labelings. It is calculated with the equation:

\begin{equation*}
\text{ARI}={R - \mathbb{E}(R) \over \max R - \mathbb{E}(R)}
\end{equation*}
where $R = (a+b)/{N \choose 2}$ is the Rand index, $a$ is the number of example
pairs that have the same class and cluster labels, $b$ is the number of pairs
that have different class and cluster labels, and $N$ is the number of examples.
The ARI score ranges from $-1$ to $1$. We define the ARI complexity measure as
$(1-\text{ARI})/2$, that maps the scores from $0$ to $1$, with $1$
indicating the highest semantic complexity.\\

\noindent \textbf{Transcript Encodings}: We used two different methods to
produce the transcript encodings: (1) the standard unsupervised method from the
field of topic modeling - Latent Dirichlet Allocation
(LDA)~\cite{blei2003latent} and (2) the contemporary sentence encoding method
from~\cite{cer2018universal}. \\

\noindent \textit{Latent Dirichlet Allocation (LDA)} is an unsupervised method
used to discover `latent' or `hidden' topics from a collection of documents
(transcripts in this case)~\cite{blei2003latent}. LDA fits a hierarchical
Bayesian model so that each transcript is represented as a distribution over
topics. The topic-distribution vectors learned by LDA for each transcript in the
dataset become the encoding used for further analysis. \\

\noindent The \textit{Universal Sentence Encoder (USE)} is a pretrained
transformer model~\cite{cer2018universal}. It outputs a 512-dimensional encoding
vector for each transcript in the dataset which we use for further
analysis. \\

\noindent Given the aforementioned methods, we compute four geometric complexity
measures for a dataset: MST-LDA, MST-USE, ARI-LDA and ARI-USE.\footnote{Note the
	proposed geometric complexity measures can be applied to any class of
	encodings beyond LDA and USE.}

\section{Experiments and Results} \label{sec:exp} Motivated by the research
questions \question{1} and \question{2} stated in section \ref{sec:intro}, we
perform our experiments in two stages.

\subsection{Quantifying Semantic Complexity} \label{sec:exp_stage1} In line with
our first research question and to draw a comparison between previously
published work and our results, we compute the semantic complexity of public SLU
datasets. We also address whether the IC accuracy reported for the STI models on
these datasets is expected in light of their complexity values.
We selected three public SLU datasets for this stage: Fluent Speech Commands
(FSC)~\cite{lugosch2019speech}, Picovoice (Pico.)~\cite{picovoice} and Snips
Smart Lights~\cite{saade2018spoken} (Snips). In the FSC and Snips datasets, the
semantic labels correspond to intents, whereas in the Pico. dataset semantic
labels were taken to be the value of the `coffeeDrink' slot since all the
examples had the same intent label (`orderDrink'). Furthermore, there are no
text transcriptions available for this dataset, so we used a speech-to-text
engine\footnote{https://aws.amazon.com/transcribe} to generate the
transcriptions.

Table~\ref{tab:public_complexity} shows the complexity values for these
datasets. We computed three n-gram entropy values: unigram, bigram and trigram
entropy. We \emph{do not} train a new STI model on these datasets and instead
note the IC accuracy reported in their original publication: FSC ($98.8\%$) and
Picovoice ($98\%$). IC accuracy with an STI model is not available for Snips;
the IC accuracy (close field: 91.72\% and far field: 83.56\%) reported
in~\cite{saade2018spoken} was obtained with a traditional, modularized SLU
design. The trend of average lexical and geometric measures among these datasets
suggest the following ascending order of semantic complexity: (1) FSC, (2) Pico.
and (3) Snips. The trend is well-connected to their respective model
performances, reflected in the the near-perfect IC accuracy numbers for FSC
($98.8\%$) and Picovoice ($98\%$) when compared to Snips (best non-STI
performance $91.72\%$). We use these numbers as a basis of comparison for our
next stage of experiments.

\begin{table}[t!]
	\centering \setlength\extrarowheight{2.5pt}
	\begin{tabular}{|l|c|c|c|c|c|}
		\hline
		\multicolumn{3}{|c|}{\textbf{Semantic Complexity Measure}} & \multicolumn{3}{c|}{\textbf{Dataset}} \\ \cline{4-6}
		\multicolumn{3}{|c|}{} & \textbf{FSC} & \textbf{Pico} & \textbf{Snips} \\ \cline{1-6}
		\toprule \hline
		\multirow{6}{*}{Lex.} & \multicolumn{2}{c|}{vocabulary} & 124 & 163 & 445 \\ \cline{2-3}
		& \multicolumn{2}{c|}{unique transcripts} & 248 & 592 & 1639 \\ \cline{2-6}
		& \multirow{4}{*}{entropy} & unigram & 5.5 & 5.5 & 6.2 \\ \cline{3-3}
		& & bigram & 7.2 & 7.3 & 9.1 \\ \cline{3-3}
		& & trigram & 7.9 & 8.8 & 10.9 \\ \cline{3-6}
		& & \textbf{average} & \textbf{6.9} & \textbf{7.2} & \textbf{8.7} \\ \hline\hline
		\multirow{6}{*}{Geo.} & \multirow{3}{*}{MST} & LDA & 0.3 & 0.8 & 0.8 \\ \cline{3-3}
		& & USE & 0.0 & 0.5 & 0.3 \\ \cline{3-6}
		& & \textbf{average} & \textbf{0.2} & \textbf{0.6} & \textbf{0.6} \\ \cline{2-6}
		& \multirow{3}{*}{ARI} & LDA & 0.02 & 0.1 & 0.4 \\ \cline{3-3}
		& & USE & 0.04 & 0.2 & 0.2 \\ \cline{3-6}
		& & \textbf{average} & \textbf{0.03} & \textbf{0.1} & \textbf{0.3} \\ 
		\hline
	\end{tabular}
	\caption{Semantic complexity measures of public SLU datasets.}
	\label{tab:public_complexity}
	\vspace*{-\baselineskip}
\end{table}

\subsection{Model Performance Versus Semantic Complexity} \label{sec:exp_stage2}
For our second research question, we want to determine if there is a correlation
between the semantic complexity of datasets and the performance of STI models
trained on these datasets.

In order to do this, we create a large, proprietary SLU dataset and apply a data
filtration scheme to generate a
sequence of SLU datasets of decreasing semantic complexity. We study the performance of two published STI models~\cite{lugosch2019speech,haghani2018audio} on these filtered datasets to analyze how strongly the model performance relates to the semantic complexity.\\

\noindent \textbf{Dataset Filtration}: We created a proprietary dataset by
extracting and annotating a window of production traffic of a commercial voice
assistant system. It contains a total of 1.6 million utterances with $>200,000$
unique transcriptions. From this original dataset, we created a sequence of
sub-datasets that have decreasing semantic complexity. The algorithm that
removes examples from a dataset to reduce the semantic complexity depends on the
particular complexity measure. For example, we reduce the n-gram entropy of a
dataset by discarding examples whose transcriptions contain an n-gram from the
least frequent set of n-grams in the dataset. We found that filtering the
dataset with this method not only decreased the n-gram entropy but
also decreased the other complexity measures. Therefore, we performed our experiments on the filtered datasets obtained using this filtering scheme.\\

\newcommand{\linebreakcell}[2][c]{%
	\begin{tabular}[#1]{@{}c@{}}#2\end{tabular}}

\begin{table*}[h!]
	\centering
	\begin{tabular}{|c|c|c|c|c|c|c|c|}
		\hline
		\multicolumn{4}{|c|}{\textbf{Semantic Complexity}} & \multicolumn{4}{c|}{\textbf{Relative Intent Classification Accuracy}} \\\hline
		\multicolumn{3}{|c|}{\textbf{Lexical Measures}} & \multirow{3}{*}{\linebreakcell{\textbf{Average}\\\textbf{MST}}} & \multicolumn{3}{c|}{\textbf{Model in~\cite{lugosch2019speech} (Pretrained)}} & \multirow{3}{*}{\linebreakcell{\textbf{Multi-task}\\\textbf{Model in {\cite{haghani2018audio}}}}} \\ \cline{1-3}\cline{5-7}
		\linebreakcell{\textbf{Average}\\\textbf{Entropy}} & \linebreakcell{\textbf{Vocab}\\\textbf{Size}} & \linebreakcell{\textbf{Unique} \\ \textbf{Transcripts}} & & \linebreakcell{No \\Unfreezing} & \linebreakcell{Unfreezing \\ Word Layer} & \linebreakcell{Unfreezing \\All Layers} & \\\hline
		11.6 & 6744 & 211585 & 0.52 & 67.5\% & 87.4\% & 94.0\% & 63.8\% \\ \hline
		7.5  & 456  & 30504  & 0.39 & 72.1\% & 91.4\% & 96.4\% & 89.0\% \\ \hline
		5.8  & 70   & 907    & 0.29 & 80.9\% & 94.8\% & 98.3\% & 94.9\% \\ \hline
		3.1  & 10   & 16     & 0.15  & 100.0\% & 100.0\% & 100.0\% & 100.0\% \\ \hline
	\end{tabular}
	\caption{Relative intent classification accuracy of STI models for datasets
		with varying semantic complexity. Each row contains the results for a
		filtered dataset. The rows are arranged in order of decreasing semantic
		complexity.}
	\label{tab:comp_res}
\end{table*}

\noindent \textbf{Model Performance}: We experiment with two popular STI model
architectures: a stacked neural model with pretrainable acoustic components
from~\cite{lugosch2019speech} and a sequence-to-sequence multi-task model
proposed by~\cite{haghani2018audio}. The acoustic component of the first model
\cite{lugosch2019speech} consists of a `phoneme layer' followed by a `word
layer', and is pretrained on the LibriSpeech
corpus~\cite{panayotov2015librispeech}. This component can be fine-tuned
following three unfreezing schedules: no unfreezing, unfreezing only the
word-layer, and unfreezing both the word and phoneme layers. The multi-task
model from~\cite{haghani2018audio} uses a shared encoder to make both transcript
and semantic predictions.

We use multiple models and training strategies in order to observe patterns that
are independent of these choices. We train both the architectures on our
filtered datasets, following their proposed training strategies and experimental
configurations. A grid search is performed on different hyper-parameter
combinations with early stopping done on the validation set. For the stacked STI
model in~\cite{lugosch2019speech}, we use the pretrained acoustic component
provided by the authors and fine-tune the model on our datasets. The multi-task
model from~\cite{haghani2018audio} was randomly initialized and trained from
scratch.

We choose the SLU task (IC) that is common to both of these models.
Table~\ref{tab:comp_res} shows the IC accuracy relative to the filtered dataset
corresponding to the last row of the Table for these models (across different
unfreezing schemes for ~\cite{lugosch2019speech}). For the sake of brevity, we
only report the results for 4 filtered datasets, note the average n-gram entropy
score (across unigram, bigram and trigram entropy) and the average MST score.
However, all observations and trends in the table were also noticed for the
average ARI measure across a total of 15 data filtration steps.

\begin{figure}[!t]
	\centering \includegraphics[width=\linewidth]{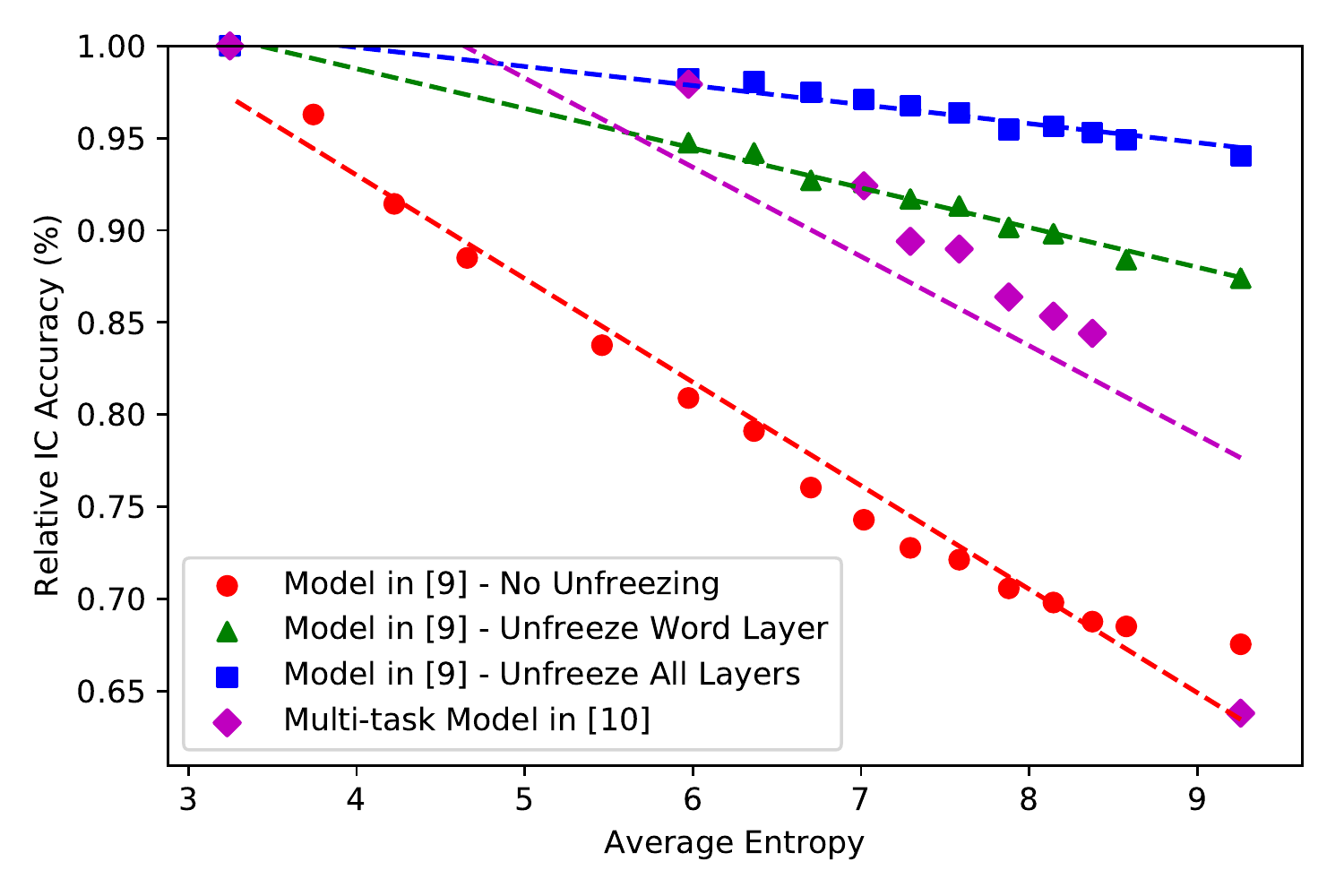}
	\caption{Relative intent classification accuracy increases as average entropy
		decreases for all model and training strategies considered.}
	\vspace{-0.7em}
	\label{fig:acc_ent}
\end{figure}

Regardless of the model architecture and unfreezing schemes, the IC accuracy of
the models consistently increases as the semantic complexity of the dataset
(both lexical and geometric) decreases. This trend holds for all the data
filtration levels, even for the ones not shown in the table.
Figure~\ref{fig:acc_ent} depicts the relationship between relative IC accuracy
and average entropy across all the filtered datasets. We notice that IC accuracy
is correlated with average entropy, depicted by the following
R\textsuperscript{2} values: model from~\cite{lugosch2019speech}: No Unfreezing:
$0.97$, Unfreezing Word Layer: $0.99$, Unfreezing All Layers: $0.94$; model
from~\cite{haghani2018audio}: $0.85$. This answers our second research question
- \textit{STI model performance increases as dataset semantic complexity
	decreases}. This trend between the computed complexity value and the model
performance validates that our proposed complexity metrics capture the inherent
semantic content of the dataset.
To determine whether the smaller sizes of filtered datasets confounded the
correlation between semantic complexity and IC accuracy, we performed a set of
experiments to remove random subsets of the original dataset. In one experiment
using the model~\cite{lugosch2019speech} with full unfreezing, removing a random
$79\%$ sample of the dataset resulted in a $2\%$ relative decrease in average
entropy and $0.03\%$ relative decrease in IC accuracy. This demonstrates that
reducing the complexity of the dataset and not just removing a random sample
accounts for the relationship depicted in Figure~\ref{fig:acc_ent}.

In line with the findings by the authors in~\cite{lugosch2019speech} on the FSC
dataset, we observe that starting from a pretrained model leads to better
overall performance than starting from a randomly initialized model (the
multi-task model~\cite{haghani2018audio} in this case). However, contrary to
their results, we observe a significant jump in accuracy if the pretrained
layers are fine-tuned. This difference is larger for datasets of higher
complexity (first row - $33\%$ difference in accuracy) than the ones with lower
complexity (last row - $10\%$ difference in accuracy). One reason we hypothesize
this happens is that with the low complexity of the dataset, especially the
public dataset FSC that has $<250$ unique transcriptions, the dense
classification layer is enough to capture that dataset specific patterns after
the acoustic component is pretrained on an external dataset. However, this
external knowledge alone is not enough to achieve comparable performance for a
complex dataset, unless the model is fine-tuned to the distinct acoustic and
linguistic patterns present in the dataset.

We found that the public datasets analyzed in Table \ref{tab:public_complexity}
had, on average, lower semantic complexity than our initial dataset. In order to
obtain average entropy values similar to the FSC, Pico, and Snips datasets,
approximately $96\%$, $92\%$, and $60\%$, respectively, of unique transcriptions
had to be removed from our initial dataset. In our own experiments, the IC
accuracies achieved on these three filtered datasets using the STI model from
\cite{lugosch2019speech} were comparable to the accuracies reported by their
respective authors. Therefore, the three public SLU datasets analyzed by us
actually fall on the lower end of the semantic complexity spectrum when compared
to the data obtained from a commercial voice assistant. Moreover, our semantic
complexity measures can indicate how well an STI architecture will perform on a
given dataset. We expect that for broader use cases, such as recognizing a rich
variety of utterance phrasings from multiple domains, the dataset semantic
complexity will be similar to the most complex dataset that we consider and the
performance of previously published models will degrade. Therefore, when
reporting STI model performance, it is important to quantify the complexity of
the task. We have shown this can be done using our proposed measures of semantic
complexity.

\section{Conclusion}
We propose several measures that can be used to quantify the semantic complexity
of the training dataset of an SLU system that employs an STI architecture. These
measures signal the difficulty of the associated SLU task. We show that our
complexity measures correlate well with STI model performance, agnostic of the
model architecture, such that as the computed complexity value decreases, the
classification accuracy increases. Our experiments reveal that the high
classification accuracy ($98-99\%$) reported on public datasets for STI models,
is associated with a low complexity value for those datasets. Our current
observations indicate that targeted use cases associated with datasets of low
complexity values are good candidates for STI architectures. However, as the use
cases get more complex, the STI performance gains diminish. Therefore, it is
important to contextualize any new STI model performance claims with the
semantic complexity values of the underlying training dataset to understand the
scope of its applicability. Notably, our complexity measures help us understand
the performance of STI models even though they are computed from transcripts,
which are not an input to the STI model. We expect that further analysis which
takes acoustic features of the speech input into account may also help us
understand the performance of STI models. In the future, we aim to utilize these
computed semantic complexity measures and representations as an additional
signal to STI models to test whether they improve SLU performance. It would also
be interesting to combine different complexity measures as a more comprehensive
metric which can generalize across all SLU tasks.

\bibliographystyle{IEEEtran}

\bibliography{mybib}

\begin{thebibliography}{10}
\providecommand{\url}[1]{#1}
\csname url@samestyle\endcsname
\providecommand{\newblock}{\relax}
\providecommand{\bibinfo}[2]{#2}
\providecommand{\BIBentrySTDinterwordspacing}{\spaceskip=0pt\relax}
\providecommand{\BIBentryALTinterwordstretchfactor}{4}
\providecommand{\BIBentryALTinterwordspacing}{\spaceskip=\fontdimen2\font plus
\BIBentryALTinterwordstretchfactor\fontdimen3\font minus
  \fontdimen4\font\relax}
\providecommand{\BIBforeignlanguage}[2]{{%
\expandafter\ifx\csname l@#1\endcsname\relax
\typeout{** WARNING: IEEEtran.bst: No hyphenation pattern has been}%
\typeout{** loaded for the language `#1'. Using the pattern for}%
\typeout{** the default language instead.}%
\else
\language=\csname l@#1\endcsname
\fi
#2}}
\providecommand{\BIBdecl}{\relax}
\BIBdecl

\bibitem{tur2011spoken}
\BIBentryALTinterwordspacing
G.~Tur, \emph{Spoken Language Understanding: Systems for Extracting Semantic
  Information from Speech}.\hskip 1em plus 0.5em minus 0.4em\relax John Wiley
  and Sons, January 2011. [Online]. Available:
  \url{https://www.microsoft.com/en-us/research/publication/spoken-language-understanding-systems-for-extracting-semantic-information-from-speech/}
\BIBentrySTDinterwordspacing

\bibitem{lee2010recent}
C.-J. Lee, S.-K. Jung, K.-D. Kim, D.-H. Lee, and G.~G.-B. Lee, ``Recent
  approaches to dialog management for spoken dialog systems,'' \emph{Journal of
  Computing Science and Engineering}, vol.~4, no.~1, pp. 1--22, 2010.

\bibitem{hinton2012deep}
G.~Hinton, L.~Deng, D.~Yu, G.~E. Dahl, A.-r. Mohamed, N.~Jaitly, A.~Senior,
  V.~Vanhoucke, P.~Nguyen, T.~N. Sainath \emph{et~al.}, ``Deep neural networks
  for acoustic modeling in speech recognition: The shared views of four
  research groups,'' \emph{IEEE Signal processing magazine}, vol.~29, no.~6,
  pp. 82--97, 2012.

\bibitem{graves2013speech}
A.~Graves, A.-r. Mohamed, and G.~Hinton, ``Speech recognition with deep
  recurrent neural networks,'' in \emph{2013 IEEE international conference on
  acoustics, speech and signal processing}.\hskip 1em plus 0.5em minus
  0.4em\relax IEEE, 2013, pp. 6645--6649.

\bibitem{bahdanau2016end}
D.~Bahdanau, J.~Chorowski, D.~Serdyuk, P.~Brakel, and Y.~Bengio, ``End-to-end
  attention-based large vocabulary speech recognition,'' in \emph{2016 IEEE
  international conference on acoustics, speech and signal processing
  (ICASSP)}.\hskip 1em plus 0.5em minus 0.4em\relax IEEE, 2016, pp. 4945--4949.

\bibitem{xu2014contextual}
P.~Xu and R.~Sarikaya, ``Contextual domain classification in spoken language
  understanding systems using recurrent neural network,'' in \emph{2014 IEEE
  International Conference on Acoustics, Speech and Signal Processing
  (ICASSP)}.\hskip 1em plus 0.5em minus 0.4em\relax IEEE, 2014, pp. 136--140.

\bibitem{ravuri2015recurrent}
S.~Ravuri and A.~Stolcke, ``Recurrent neural network and lstm models for
  lexical utterance classification,'' in \emph{Sixteenth Annual Conference of
  the International Speech Communication Association}, 2015.

\bibitem{sarikaya2014application}
R.~Sarikaya, G.~E. Hinton, and A.~Deoras, ``Application of deep belief networks
  for natural language understanding,'' \emph{IEEE/ACM Transactions on Audio,
  Speech, and Language Processing}, vol.~22, no.~4, pp. 778--784, 2014.

\bibitem{lugosch2019speech}
L.~Lugosch, M.~Ravanelli, P.~Ignoto, V.~S. Tomar, and Y.~Bengio, ``Speech model
  pre-training for end-to-end spoken language understanding,'' \emph{arXiv
  preprint arXiv:1904.03670}, 2019.

\bibitem{haghani2018audio}
P.~Haghani, A.~Narayanan, M.~Bacchiani, G.~Chuang, N.~Gaur, P.~Moreno,
  R.~Prabhavalkar, Z.~Qu, and A.~Waters, ``From audio to semantics: Approaches
  to end-to-end spoken language understanding,'' in \emph{2018 IEEE Spoken
  Language Technology Workshop (SLT)}.\hskip 1em plus 0.5em minus 0.4em\relax
  IEEE, 2018, pp. 720--726.

\bibitem{chen2018spoken}
Y.-P. Chen, R.~Price, and S.~Bangalore, ``Spoken language understanding without
  speech recognition,'' in \emph{2018 IEEE International Conference on
  Acoustics, Speech and Signal Processing (ICASSP)}.\hskip 1em plus 0.5em minus
  0.4em\relax IEEE, 2018, pp. 6189--6193.

\bibitem{serdyuk2018towards}
D.~Serdyuk, Y.~Wang, C.~Fuegen, A.~Kumar, B.~Liu, and Y.~Bengio, ``Towards
  end-to-end spoken language understanding,'' in \emph{2018 IEEE International
  Conference on Acoustics, Speech and Signal Processing (ICASSP)}.\hskip 1em
  plus 0.5em minus 0.4em\relax IEEE, 2018, pp. 5754--5758.

\bibitem{hubert1985comparing}
L.~Hubert and P.~Arabie, ``Comparing partitions,'' \emph{Journal of
  classification}, vol.~2, no.~1, pp. 193--218, 1985.

\bibitem{blei2003latent}
D.~M. Blei, A.~Y. Ng, and M.~I. Jordan, ``Latent dirichlet allocation,''
  \emph{Journal of machine Learning research}, vol.~3, no. Jan, pp. 993--1022,
  2003.

\bibitem{cer2018universal}
D.~Cer, Y.~Yang, S.-y. Kong, N.~Hua, N.~Limtiaco, R.~S. John, N.~Constant,
  M.~Guajardo-Cespedes, S.~Yuan, C.~Tar \emph{et~al.}, ``Universal sentence
  encoder,'' \emph{arXiv preprint arXiv:1803.11175}, 2018.

\bibitem{picovoice}
Picovoice, ``Speech-to-intent benchmark,''
  \url{https://github.com/Picovoice/speech-to-intent-benchmark}, 2020.

\bibitem{saade2018spoken}
A.~Saade, A.~Coucke, A.~Caulier, J.~Dureau, A.~Ball, T.~Bluche, D.~Leroy,
  C.~Doumouro, T.~Gisselbrecht, F.~Caltagirone \emph{et~al.}, ``Spoken language
  understanding on the edge,'' \emph{arXiv preprint arXiv:1810.12735}, 2018.

\bibitem{panayotov2015librispeech}
V.~Panayotov, G.~Chen, D.~Povey, and S.~Khudanpur, ``Librispeech: an asr corpus
  based on public domain audio books,'' in \emph{2015 IEEE International
  Conference on Acoustics, Speech and Signal Processing (ICASSP)}.\hskip 1em
  plus 0.5em minus 0.4em\relax IEEE, 2015, pp. 5206--5210.

\end{thebibliography}

\end{document}